\documentclass[5p]{elsarticle}
\usepackage{hyperref}
\usepackage[dvipsnames]{xcolor}
\usepackage{soul}

\journal{a Journal}

\usepackage{amsfonts}
\usepackage{float}
\usepackage{amsmath}
\usepackage{multirow}

\begin{document}

\begin{frontmatter}

\title{Latent regularization for feature selection using kernel methods in tumor classification}




\author[mymainaddress,mysecondaryaddress,mythirdaddress]{Martin Palazzo\corref{mycorrespondingauthor}}
\cortext[mycorrespondingauthor]{Corresponding author}
\ead{martin.palazzo@utt.fr}
\author[mysecondaryaddress]{Patricio Yankilevich}
\author[mymainaddress]{Pierre Beauseroy}


\address[mymainaddress]{Institut Charles Delaunay, Universite de Technologie de Troyes}
\address[mysecondaryaddress]{Biomedicine Research Institute of Buenos Aires - Max Planck Partner Institute}
\address[mythirdaddress]{Universidad Tecnologica Nacional Buenos Aires}

\begin{abstract}
The transcriptomics of cancer tumors are characterized with tens of thousands of gene expression features. Patient prognosis or tumor stage can be assessed by machine learning techniques like supervised classification tasks given a gene expression profile. Feature selection is a useful approach to select the key genes which helps to classify tumors. In this work we propose a feature selection method based on Multiple Kernel Learning that results in a reduced subset of genes and a custom kernel that improves the classification performance when used in support vector classification. During the feature selection process this method performs a novel latent regularisation by relaxing the supervised target problem by introducing unsupervised structure obtained from the latent space learned by a non linear dimensionality reduction model. An improvement of the generalization capacity is obtained and assessed by the tumor classification performance on new unseen test samples when the classifier is trained with the features selected by the proposed method in comparison with other supervised feature selection approaches.
\end{abstract}


\end{frontmatter}


\section{Introduction}

Computational biology community is producing large quantities of raw data from different organisms. More precisely, cancer systems are being recorded at a high pace at big repositories like The Cancer Genome Atlas (TCGA) \cite{weinstein2013cancer} and the International Cancer Genome Consortium (ICGC) \cite{international2010international} data portals. These databases contains hundred of tumor records from humans, each one with data from different omics like Genomics, Transcriptomics and Proteomics among others. Furthermore these records are related with the clinical data of the patient, like survival time and tumor stage. Transcriptomics or gene expression is considered as an important layer of information to be used as predictor for clinical outcomes of the patient \cite{beer2002gene}. Gene expression data can be used to predict patient prognosis or tumor stage by modeling the problem as a binary classification one. Nevertheless, there are over 20.000 protein coding genes and this high dimensional context increases the complexity of the prediction problem. For this reason it is necessary to reduce the initial high dimensional problem while keeping the interpretability of the features involved. This approach should determine which genes are more important to predict the clinical outcome and involves not only a classification problem but also a feature selection one.\\ 
Machine learning is a powerful approach to tackle these two problems by using algorithms that learns rules from data. Supervised feature selection is an important field dedicated mainly to select a subset of features that improves a classification task with the assumption that the initial feature set involves not only useful predictors but also noisy features that should not be considered. More specificaly, tumor profiles are described by the expression of thousands of genes but the majority of these do not provide useful signals to achieve a classification task for tumor stage or patient prognosis. Moreover, in general a reduced subset of $p$ genes from an initial feature set of $n$ genes is just necessary to classify between tumor profiles with an acceptable performance where $p<<n$. Therefore Feature Selection methods are used in biomarker discovery \cite{he2010stable}. Most of the time, the important features are selected using a supervised objective function \cite{li2017feature}. In some cases this supervised objective may be too strict and difficult to fulfill in order to obtain a model that could generalize on new unseen data \cite{reunanen2003overfitting}. Then a major question arises: is it possible to improve the feature selection process by taking advantage of the structure in feature space of training data in the search for better classification and generalization performances?\\ 
Our work proposes a feature selection method based on Multiple Kernel Learning (MKL) \cite{gonen2011multiple}, a family of models which defines a new kernel by combining multiple kernel functions via a weighting system to optimize an objective function. Additionally the proposed method combines MKL and a nonlinear latent feature extraction model to improve the feature selection problem by a combination of supervised and unsupervised approaches respectively. This combination of approaches aims to improve the generalization capacity in classification of the selected features. This strategy aims to maximize the separability between tumor classes while considering simultaneously the latent structure of the training data. In this work the proposed selection method performs what we name a latent regularization using simultaneously the labels of the data and unsupervised latent variables to improve the classification task and the generalization. To extract the latent variables of the training data an unsupervised dimensionality reduction has been performed using kernel-PCA (kPCA) \cite{mika1999kernel}. With our method the selected features are the ones which consider the tumor labels and the data structure at the same time. \\
The proposed method is designed to deal with tumor classification problems where dimensionality $n > 18.000$ and the sample size is lower than 200 tumor profiles. Tumor profiles are classified by stage or prognosis. In this scenario most of feature selection algorithms may under perform due to the lack of tumor samples and the high dimensional feature set.  
The combination of both MKL and kPCA models perform feature selection while improving the generalization capacity of the classification task due to the consideration of the latent variables extracted by the kPCA. This latent regularisation step works as a label relaxation that is shown to improve the tumor classification performance on new unseen test samples. The proposed method is named Kernel Latent Regularization Feature Selection (KLR-FS).\\


\section{Related Work}

Feature selection is a key task in biology due to the high number of genes describing a tumor profile. Feature selection algorithms are useful in this type of problems and are categorized in filter, wrapper and embedded \cite{ang2015supervised}. Methods like LASSO \cite{algamal2015penalized}\cite{moon2016stable}, Minimal Redundancy Maximal Relevance (MRMR) \cite{zhang2016improve}, Recursive Feature Elimination (RFE) \cite{duan2005multiple} and Hilbert Schmidt Independence Criterion Kernelized coupled with Lasso (HSIC-Lasso) \cite{yamada2014high} have been used to select key genes on expression data for tumor classification.\\
Supervised MKL \cite{gonen2011multiple} relies on a linear combination of kernels to improve the classification performance of a support vector machine classifier \cite{rakotomamonjy2008simplemkl} or to maximize the Kernel Target Alignment (KTA) score \cite{pothin2006greedy}\cite{aiolli2015easymkl} of the resulting kernel. Different implementations of supervised MKL have been used in cancer problems for patient prognosis prediction. MKL has been applied for multi-omics data fusion considering gene expression, copy number variation, gene methylation, mRNA expression and histopathological features \cite{zhang2019novel} for Glioblastoma Multiforme survival prediction while optimizing the MKL problem to improve a Support Vector Machine (SVM) classification task. The same approach of multi-omics fusion using MKL has been used in 14 types of cancer for prognosis assessment  by combining the kernels in order to improve the Kernel Target Alignment (KTA) score \cite{zhu2017integrating} \cite{cristianini2002kernel}. Also the use of MKL has been used to combine multiple feature subsets by optimizing the MKL model with the classification performance of a SVM model in Lung Squamus cell carcinoma samples \cite{zhang2019lscdfs}. Moreover, MKL has been used to select subsets of gene features using gene expression data from 9 cancer types where the MKL solution is optimized to improve a SVM classifier and where the result defines the feature subset that optimizes the classification accuracy \cite{du2017feature}. \\
Unsupervised dimensionality reduction approach to perform non linear extraction of latent features have been widely used in cancer problems involving molecular data \cite{palazzo2019pan} \cite{liu2005gene}. kPCA is the nonlinear dimensionality reduction method used in this work. It has been used on gene expression data for missing data estimation \cite{shan2009kernel}, to integrate biological data from different sources of a nutrigenomic study in mouse \cite{reverter2014kernel}, to find meaningful representations from Leukemia and Lymphoma microarray data \cite{reverter2012kernel} and applied on Breast cancer tumor profiles for multi-omic data fusion \cite{mariette2018unsupervised}. \\ 
There is not doubt about the potential of kernel-PCA, Multiple kernel learning and feature selection approaches on cancer genomics. The high dimensional context of the gene expression data makes these methods useful to reduce the dimensionality and to perform biomarker discovery via feature selection. Moreover, we observe that most of the supervised feature selection approaches do not take advantage of the structure of the training data \cite{ang2015supervised}. From our best knowledge, there is not any prior work that mix via kernel methods the latent data structure and the class labels to improve the classification performance of a supervised feature selection method.

\section{The KLR-FS method}

As introduced in the previous section, this work proposes a feature selection method that aims to retain information about labels and about the global data structure. To achieve that goal a new feature selection method is proposed which uses a regularization step that mixes target sample labels with unsupervised latent variables learned from the training samples. This section presents the three main parts of the KLR-FS method. First the feature selection strategy using MKL maximizing the kernel alignment to select a subset of genes by a classic supervised criteria where labels are used as the only target. Then the latent regularisation is introduced by explaining how to build a target kernel matrix composed by a mix of both supervised labels and unsupervised latent variables in order to select features that considers both labels and structure. Finally support vector classification is performed on tumor profiles using the resulting selected features and the kernel obtained from the MKL step. 

\subsection{Feature selection with multiple kernel learning}
By a Multiple Kernel Learning process built to optimize the KTA score a subset of features is selected and a kernel is obtained. This subsection defines how to select features with MKL by a supervised approach. \\
Given \(\mathcal{X} \subseteq \mathbb{R}^{n}\) a n-dimensional space , a function  \(k:\mathcal{X}\times \mathcal{X} \rightarrow \mathbb{R}\) which is symmetric and semi-positive definite meets the condition
\begin{equation}
    k\left ( x_i, x_j \right ) = \left \langle \phi (x_i) , \phi (x_j)  \right \rangle_H
\end{equation}
where \(\phi\) is a mapping function from \(\mathcal{X}\) to a high dimensional feature Hilbert space \(\mathcal{H}\)  such that
\begin{equation}
\phi : \mathcal{X} \mapsto \phi \left ( \mathcal{X} \right ) \in \mathcal{H}
\end{equation}
and $\mathcal{H}$ is defined as the Reproducing Kernel Hilbert Space (RKHS) of $k$ \cite{vert2004primer}. The function $k$ computes the inner product of a pair of vectors in $\mathcal{H}$ the Hilbert Space.
Considering a set of $m$ labeled samples such that $$\left \{ \left ( x_1,y_1 \right ),\left ( x_2,y_2 \right ), ...,\left ( x_m, y_m \right ) \right \}$$
where $x_i \in  \mathcal{X}$ and $y_i \in \left \{ -1,+1 \right \}$ the Kernel Matrix or Gram Matrix $G$ is defined as a $M \times M$ matrix with entries $K_{ij}$. Every entry of the Kernel or Gram Matrix is defined as
\begin{equation}
K_{ij} = \left \langle \phi \left ( x_i \right ), \phi \left ( x_j \right ) \right \rangle = k\left ( x_i, x_j \right )
\end{equation}
The kernel function can be thought as an information bottleneck that concentrates the information required to perform a learning task \cite{shawe2004kernel}. Kernels can be thought as similarity functions where an output close to $0$ means orthogonal samples or high dissimilar and an output close to $1$ means similar samples in the Hilbert space.  In this work we will use the Radial Basis Function (RBF) Kernel which is defined as
\begin{equation}
k(x_i, x_j) = \textup{exp}\left ( - \gamma \left \| x_i-x_j \right \|^{2} \right )
\end{equation}
where $\gamma$ is a scale parameter. \\
Given two valid kernels $K_1$ and $K_2$ over a set of samples $M$, the alignment $A$ between both kernels is defined as
\begin{equation}
\mathit{A}\left ( K_1, K_2 \right ) = \frac{\left \langle K_1, K_2 \right  \rangle_F }{\sqrt{\left \langle K_1, K_1 \right  \rangle_F \left \langle K_2, K_2 \right  \rangle_F}}
\end{equation}
and measures the similarity between the two kernels using the same sample set $M$ \cite{cristianini2002kernel}\cite{kandola2002extensions}. Here the operation $\left \langle . \right  \rangle_F$ is the Frobenious Inner product. The alignment can be thought as how similar both kernels maps the samples from $M$. If tumor labels are used, then $K_2$ represents an ideal Kernel matrix or target $K_{yy}$ where $K_{yy} = +1$ if  $y_i=y_j$ and $K_{yy} = 0$ if $y_i \neq y_j$. The alignment of a kernel $K$ built on $x$ and the target $K_{yy}$ can be expressed as

\begin{equation}
\mathit{A}\left ( K, K_{yy} \right ) = \frac{\left \langle K, K_{yy} \right  \rangle_F }{\sqrt{\left \langle K, K \right  \rangle_F \left \langle K_{yy}, K_{yy} \right  \rangle_F}}
\end{equation}

known as Kernel Target Alignment (KTA) score. The higher the KTA between a given kernel matrix $K$ and its target $K_{yy}$ the higher the inter-cluster separation between the two classes. \\
Multiple Kernel Learning (MKL) allows the combination of simple kernels to build a more complex one with higher KTA. MKL is defined as the linear combination of multiple kernels to build a final one \cite{gonen2011multiple} and can be expressed as

\begin{equation}
\boldsymbol{K_{\mu }}\left ( \boldsymbol{x},\boldsymbol{x{}'} \right ) = \sum_{i=1}^{n} \mu _i K_i\left ( \boldsymbol{x},\boldsymbol{x{}'} \right ), \mu _i \geq 0
\end{equation}

where the vector parameter $\mu$ corresponds to the weight $\mu_i > 0$ of each kernel $k_i$ and it is directly related to the importance of each kernel in the final solution. Despite there are different objective functions to calculate the weights of the MKL model, like the classification accuracy of a support vector machine, in this work the MKL model is built with the objective to maximize the KTA. This means that the resulting kernel from the combination of the initial ones will present a higher inter-class separability than each individual kernel on its own. The resulting KTA for the kernel $\boldsymbol{K_{\mu }}$ is $\mathit{A}\left ( \boldsymbol{K_{\mu }}, K_{yy} \right )$ and computed as detailed in equation 6.\\
This work uses a greedy strategy \cite{pothin2006greedy} to compute the weights $\mu$ of the resulting  $\boldsymbol{K_{\mu }}$ kernel by combining just two kernels at each iteration while maximizing the KTA of the resulting kernel. This strategy allows a simpler computation of the $\mu$ vector by obtaining the weights values where the derivative of the alignment becomes $=0$ as optimal condition for each partial derivative. Solving the MKL process with two kernels at each iteration $t$ means that the weight vector is $\boldsymbol{\mu} = \left [ \mu_{\alpha},\mu_{\beta}  \right ]$. Since the solution with two kernels is convex \cite{cortes2012algorithms} the values of $\mu_{\alpha}$ and $\mu_{\beta}$ can be obtained from 
\begin{equation}
\begin{matrix}
\frac{\partial \boldsymbol{KTA}\left ( \mu_{\alpha},\mu_{\beta} \right )}{\partial \mu_{\alpha}} =0 & \frac{\partial \boldsymbol{KTA}\left ( \mu_{\alpha},\mu_{\beta} \right )}{\partial \mu_{\beta}} =0 
\end{matrix}
\end{equation}
subject to $\mu_{\alpha},\mu_{\beta} \geq  0$. From a set of $N$ kernels, the greedy approach starts at iteration $t = 0$ by choosing the kernel $K_i$ from $N$ with highest $KTA_i$. Then assign $K_i$ to a empty list $P$ containing the kernels which are part of the solution and makes it the current solution $K_{\mu}^{(t)} = K_i$. Then for the following iterations $t = 1...p$ the method combines iterativelly the current $K_{\mu}^{(t)}$ and a new kernel $K_j$ from $N$ with $i \neq j$ to obtain a new solution $ K_{\mu}^{(t+1)} = \mu_{\alpha} K_{\mu}^{(t)} +  \mu_{\beta} K_j$  that maximizes the overall KTA as $A(K_{yy},K_{\mu}^{(t+1)})$. This process prioritizes in adding first to the solution $P$ the kernels that increase the most the overall KTA, thus the kernel selected first will have the highest weight $\mu_i$. The following kernels to be selected will have a lower weight $\mu_i$ since their contribution to maximize the overall KTA as $\Delta KTA$ is going to be lower than the previous added kernel. The KTA increment at each iteration will be $\Delta KTA^{(t)} > \Delta KTA^{(t+1)}... >\Delta KTA^{(p)}$. This means that when a number $p$ of kernels is requested to be in the solution weights tend to decrease with $p$ as $\mu_0 > \mu_1 > .... > \mu_p$.\\
Given a dataset of $m$ samples characterized by $n$ features, this work proposes to build $n$ feature-wise kernels $K_i$. Feature-wise kernel $\gamma_i$ parameters are selected based on the criterion of maximizing the alignment of each kernel. Then using the proposed MKL method a subset of feature-wise kernels $P$ is iterativelly selected and combined to increase the overall KTA of the resulting kernel $\boldsymbol{K}_{\mu}$. Only the feature-wise kernels that increases the KTA are included in the final kernel. This approach leads to a sparse solution where the number of selected features $p$ associated to the selected feature-wise kernels is $p << n$. 
The desired output of the greedy MKL approach is a reduced set of $p$ features and a kernel $\boldsymbol{K}_{\mu}$ that improve the inter-cluster distance between samples of different tumor classes and thus improve the support vector classification. The positive values of the weighting vector $\mu_i > 0$ indicates the feature importance on the result and the selected features.\\
Once $\boldsymbol{K}_{\mu}$ is built it is used as a custom kernel function in a support vector classifier \cite{vert2004primer} for binary classification in tumor stage and patient survival problems.

\subsection{Latent regularization with nonlinear feature extraction}

In the previous section a feature selection method based on multiple kernel learning and kernel target alignment is described. This method selects features using a supervised kernel matrix $K_{yy}$ where the separation between classes is ideal as target. But this objective is very specific and since the number of samples is small it can lead to significant different solutions when changing samples in the training set. Thus it may be interesting to be less specific and more conservative about the structure of the data. For this reason a relaxation of this supervised constraint is proposed in this work by mixing the target kernel $K_{yy}$ with a $K_{z}$ kernel built on the latent space $Z$ learned by a dimensionality reduction algorithm based on a non-linear transformation $\phi_z(x)$. Note that the $K_{yy}$ is built from the tumor labels and the $K_{z}$ from the extracted latent variables, thus the first one is built by a supervised approach and the latter by an unsupervised one. The mixture of both kernels forms a new target kernel $K_{\delta}$ that has supervised and unsupervised information from the training samples. For the unsupervised and nonlinear dimensionality reduction transformation $\phi_z(x)$ any algorithm can be used such as Kernel-PCA, Autoencoders or t-SNE. The key aspect is to learn using an unsupervised criteria, a latent space from the training samples to capture the structure of the data and to mix it with the supervised labels as a mean to relax the supervised constraint.\\ 
Most supervised feature selection models $f$ are expressed as
$$
y \sim f(x)
$$
where a subset $p$ of selected features minimizes a loss function between the model output $f(x)$ and the true labels $y$. This work proposes a feature selection model that learns not only from the true labels $y$ but also from the latent structure of the training data as
$$
(z,y) \sim f(x)
$$
where $z = \phi(x)_z$ is the latent space obtained from a nonlinear and unsupervised mapping function. The mixture between the labels $y$ and $z$ creates a hybrid target composed by a supervised and unsupervised approach. In this work to extract a latent space $z$ the function $\phi(x)_z$ is learned by a Kernel-PCA method \cite{mika1999kernel}. \\
This method is based on a nonlinear mapping $\phi_z(x)$ from $\mathcal{X}$ to $\mathcal{H}$ where the standard Principal Component Analysis (PCA) algorithm is performed.\\
The PCA is a linear dimensionality reduction method and is defined as the orthogonal projection of the training samples into a low dimensional space such that the variance of the projected samples is maximized \cite{bishop2006pattern}. From a set of samples ${x_m}$ where $m = 1, ..., M$ characterized in $n$ dimensions, the goal of PCA is to map $x_m$ into a low dimensional space of dimension $p$ such that $p < n$. The PCA projects the data into the p low dimensional space that maximizes the variance. The PCA transformation is obtained by applying the spectral decomposition to the covariance matrix $C$ of the training data $x_m$ such that $U \Lambda = C U$. The $U$ matrix contains stacked vectors $u_1, ..., u_p$ where $u_i$ is the \textit{ith} eigenvector corresponding to the \textit{ith} largest eigenvalue $\lambda_i$ in the $\Lambda$ matrix. Then by selecting the first $p$ eigen-vectors a new $n \times p$ matrix ${U}'$ is obtained and works as the desired subspace. Then by computing $X{U}' = Z$ the original data is projected by ${U}'$ into $Z$ of dimension $p$.\\
To make PCA nonlinear the Kernel Methods introduced in the previous section are used to define the Kernel-PCA in order to perform an implicit PCA in the projected Hilbert space $\phi(x)$ via the kernel trick \cite{scholkopf2001kernel}. In this case the spectral decomposition is applied to a Kernel Matrix $K$ such that 
$$m \Lambda U = K U$$ 
where $U$ and $\Lambda$ are the matrices containing the eigenvectors and eigenvalues respectively and $m$ the number of samples. Then the resulting coordinates $(l_1, ..., l_p)$ known as kernel principal components are calculated by 
\begin{equation}
\begin{matrix}
l_j = \sum_{i=1}^{m} u_{ij} k(x_i,x) & ,j = 1...p
\end{matrix}
\end{equation}
and this is equivalent to perform a nonlinear PCA in the original input space.\\
Once the kernel-PCA is computed with the training samples $\boldsymbol{x}_{tr}$ these are mapped to $\boldsymbol{z}$ and a kernel $k_{z}$ is built on the latent space. $k_{z}$ captures the unsupervised structure of the training samples in the latent space. Then by a linear combination of the supervised $k_{yy}$ and unsupervised $k_{z}$ kernels  a new hybrid target kernel $K_{\delta}$ can be created
\begin{equation}
K_{\delta} = \delta K_{yy} + (1-\delta) K_{z}
\end{equation}
This new kernel contains a linear combination of both the supervised labels and unsupervised latent variables ruled by a new parameter named as \textit{Mixture Coefficient} $\delta \in [0,1]$. 

Figure 1 shows the KLR-FS pipeline. From equation 10 three scenarios are possible. When $\delta = 1$ then $K_{\delta} = K_{yy}$ and corresponds to the supervised kernel described in the previous section and the feature selection process is completely supervised. On the other hand when $\delta = 0$ then $K_{\delta} = K_{z}$ and it represents the unsupervised structure provided by the kernel-PCA of the training samples. When $\delta = 0$ the target $K_{\delta}$ does not contain any supervised information and the MKL process is unsupervised. Finally, every value of $0<\delta<1$ corresponds to a kernel $K_{\delta}$ that has a mixture of supervised and unsupervised components of the problem. Then by this approach a latent regularization of the supervised feature selection problem is possible. The hypothesis stated in this work is based on the assumption that the features selected by MKL in a mixture of supervised labels and unsupervised latent variables results in a higher generalization ability and thus in higher classification performance. \\
Finally, once the MKL step is done a $\boldsymbol{K}_{\mu}$ kernel and a subset $P$ of selected features are obtained as result. Then the $\boldsymbol{K}_{\mu}$ kernel is used as the kernel in a support vector classification \cite{scholkopf2002learning} to classify tumor profiles. 
$$
\hat{y}(\mathbf{x}) = \sum_{i=1}^{M}\alpha _i y_i \left \langle \phi_{\mu} \left ( x \right ), \phi_{\mu} \left ( x_i \right ) \right \rangle_{\mathcal{H}} + b = \sum_{i=1}^M \alpha_i y_i k_{\mu}( \mathbf{x},  \mathbf{x}_i) +b
$$

\begin{figure*}[h!]
  \centering 
  \includegraphics[width=\textwidth]{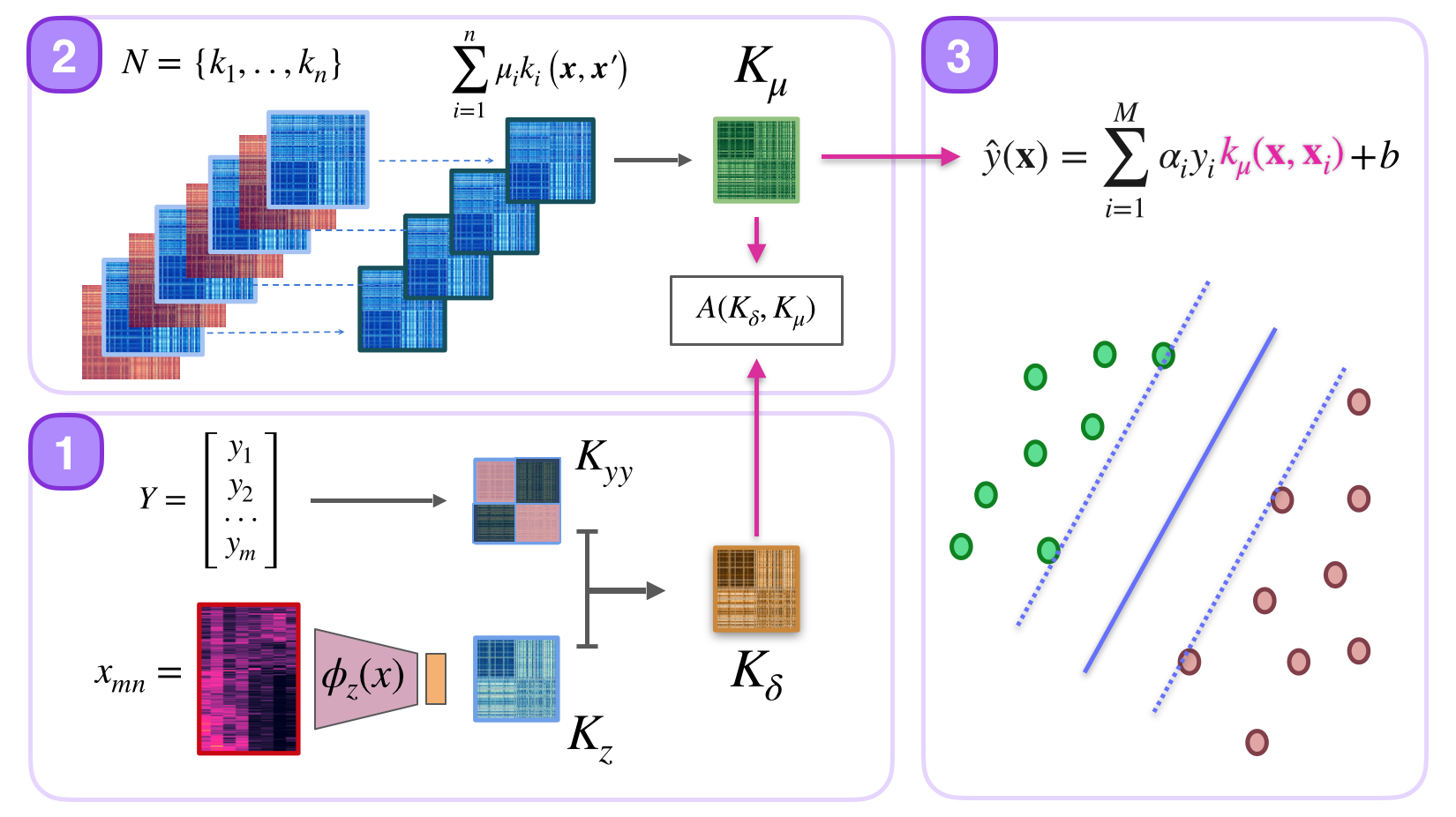} 
  \vspace*{-4mm}
  \caption{The KLR-FS pipeline. 1) A $K_{yy}$ matrix is built using the tumor labels. A kernel-PCA model is trained using the training data and a kernel $K_{z}$ is built on the latent space learned from the kernel-PCA. The $K_{\delta}$ kernel is obtained from the mixture between $K_{yy}$ and $K_{z}$. 2) From the training data a set $N$ of feature-wise kernels is built. By MKL a subset of feature-wise kernels are selected and a $K_{\mu}$ kernel is obtained by improving the alingment with $K_{\delta}$ kernel. The $\mu$ vector indicates the selected features.  3) The $K_{\mu}$ kernel is used in Support Vector Classification.}
  \label{fig:example} 
\end{figure*} 

\subsection{Model evaluation}

This work is focused on binary classification of tumor samples by using reduced feature subsets selected by KLR-FS. The classification task gives an idea about the predictive power of the selected features. Each binary classification problem has two classes: a Positive and a Negative class. To evaluate a classification model on new independent test samples four outcomes are possible: True Positive (TP), False Negative (FN), True Negative (TN) and False Positive (FP). Given the Positive class the TP counts the number of samples well classified while the FN counts the wrong classified by the model. Given the Negative class the TN counts the number of samples well classified while the FP counts the wrong classified ones. \\
Another way to measure the performance of a classifier is to estimate the Area Under the Receiving Operation Characteristic Curve (ROC) \cite{melo2013area} or Area Under the Curve (AUC). The ROC represents the TP rate as a function of the FP rate. The AUC is a score used to measure the overall performance of a binary classifier. It ranges from $0.5$ to $1$. An AUC $= 0.5$ represents the performance of a random classifier while an AUC $= 1$ corresponds to a perfect classifier. Unlike Accuracy or other classification metrics based on a specific decision threshold defined a priori by the classifier the AUC is independent of the classification threshold and is a measure of a continuous output. This makes the AUC a robust score. The AUC is used to evaluate the classification performance of a binary classifier trained on the selected features by KLR-FS. For all the experiments a Support Vector Classifier is trained using the selected features of each method. \\
To evaluate the robustness and quality of the feature selection process the Redundancy Rate (RED) \cite{zhao2010efficient} \cite{yamada2014high} metric is used. The RED score measures the mean value of absolute correlation between features is computed as
\begin{equation}
\text{RED} = \frac{1}{p(p-1)}\sum_{f_i, f_j \in \textbf{P}} |\rho_{ij}|
\end{equation}

where $\rho_{ij}$ is the correlation score between the \textit{ith} and the \textit{jth} selected variables. The RED metric takes values between $0$ and $1$. A low value of RED indicates that the majority of the selected features within the subset $P$ have low linear correlation between each other and thus a low redundancy is expected within the selected feature set which can be interpreted as a high quality of selection. On the other side values of RED close to $1$ correponds to a subset of features with high redundancy which is a non desired output.




\subsection{Benchmark methods}
The proposed feature selection method for classification is compared with four other supervised feature selection methods. These methods are  Minimum Redundancy Maximum Relevance (mRMR)\cite{peng2005feature}, the High-Dimensional Feature Selection by Feature-Wise Kernelized Lasso (HSIC-Lasso) \cite{yamada2014high}, the SVM Recursive Feature Elimination (SVM-RFE)\cite{duan2005multiple} and the univariate ANOVA filter \cite{lazar2012survey}.\\ 
The mRMR method attempts to discard redundant features while keeping the features with highest relevance to the target labels $y$. The objective is to select a feature subset that best characterizes the statistical property of a target label \cite{peng2005feature}. The selection has the constraint that selected features are mutually as dissimilar to each other as possible, but marginally as similar to the target label.\\
The HSIC-Lasso method \cite{yamada2014high} is a feature-wise kernelized Lasso that captures non-linear dependency between input features and target labels.\\
The SVM-RFE is a greedy feature selection method \cite{adorada2018support}\cite{guyon2002gene} that generates a ranking list of features and selects a subset of the top-ranked features. The ranking is built by a feature weight vector $w$ obtained from the parameters of the hyperplane decision function of a SVM classifier and the top $p$ features are selected.\\ 
Finally, ANOVA t-test is an univariate feature selection method that gives a score to each feature based on the \textit{p-value} obtained from a statistical test between the feature $i$ and the class label $y$.\\
During support vector classification for benchmark methods a RBF Kernel has been used with a grid search on gamma hyperparameter $\gamma = [0.01,0.1,1,10]$. For the KLR-FS the resulting $k_{\mu}$ kernel is used. A grid search for the C hyperparameter $C = [0.1,1,10,100]$ has been used for all kernels. Hyperparameters have been selected by 5-fold cross validation on train set to select the best model. 

\section{Datasets}

In this work three cancer datasets containing tumor profiles characterized by gene expression (RNA-Seq) features are used to evaluate the proposed method. These datasets are the Breast Cancer BRCA-US, the Pancreas Cancer PACA-CA and the Lung Cancer SMK-CAN-187. The BRCA-US and PACA-CA datasets are available from the International Cancer Genome Consortium \cite{international2010international}. The SMK-CAN-187 has been presented by \cite{spira2007airway} and available from \url{http://featureselection.asu.edu/} and the Gene Expression Omnibus \cite{edgar2002gene} under GEO accession number GSE4115. \\ 
The Breast cancer BRCA-US dataset is composed by 194 Breast cancer samples labeled by the survival days since diagnosis. To define two classes the threshold between low and high survival is defined as five years of survival since diagnosis \cite{chen2014trends}. The BRCA-US tumor samples are characterized by the expression of 20502 protein coding genes.\\
The Pancreas cancer PACA-CA dataset is composed by 135 tumor profiles labeled by tumor stage. The stage labels are IA, IB, IIA and IIB. The stages IA and IB are considered early stage while the stages IIA and IIB are considered late stage. The PACA-CA tumor samples are characterized by the expression of 18020 protein coding genes. \\
The SMK-CAN-187 is a benchmark microarray based gene expression database and it has 187 samples and 19993 gene expression features.

\begin{table}[H]
  \centering 
 \begin{tabular}{|c |c| c| c|} 
 \hline
 Dataset & Gene features & Class & Samples \\ [0.5ex] 
 \hline\hline
 \multirow{2}{*}{BRCA-US} & \multirow{2}{*}{20502} & Low Survival & 63 \\ 
 \cline{3-4}
 &  & High Survival & 131 \\
 \hline
\hline
  \multirow{2}{*}{PACA-CA} & \multirow{2}{*}{18020} & Early Stage & 84 \\
 \cline{3-4}
  &  & Late Stage & 51 \\ [1ex] 
 \hline
 \hline
  \multirow{2}{*}{SMK-CAN-187} & \multirow{2}{*}{19993} & Control & 90 \\
 \cline{3-4}
  &  & Tumor & 97 \\ [1ex] 
 \hline
\end{tabular}
\caption{Size of each dataset.}
\label{tab:example}     
\end{table}

Table 1 summarizes the size of each data set. It is clearly visible the high dimensional context of the three datasets and the low number of tumor samples.

\subsection{Preprocessing}

The initial sample set has been randomly split between $80\%$ as train and $20\%$ as test. By using just the training samples all the gene expression features have been auto-scaled  with 0 mean and unit variance. Test samples have been auto-scaled using the transformation learned from train samples. Then by using training samples the feature selection methods are applied and a subset of features are selected. With the selected features a support vector classifier is trained and tuned by 5-fold cross validation within train set for hyperparameter selection and evaluated on the test set. 

\section{Experimental Results}

For performance estimation the split between train and test set has been done randomly five times and at each time a feature selection and classification tasks are implemented. Then the classification results and the redundancy rate of the feature selection are averaged across all random splits and the mean and standard deviation of these metrics are reported. \\
The experimental results section is divided in two subsections. The first one is devoted to the Latent Regularization for feature selection and analyzes how the performance of KLR-FS behaves for different values of the mixture coefficient described in equation 10. The second subsection is the performance estimation and statistical comparison of the proposed method and the benchmark ones.

\subsection{Latent regularization for feature selection}
This subsection details how the feature selection process of KLR-FS behaves with different settings of the mixture coefficient $\delta$. As explained in the materials and methods section by varying the values of $\delta$ different target kernels $K_{\delta}$ are obtained, each one with a different mix between the unsupervised $K_z$ and the supervised $K_{yy}$ kernels. A set of six values of the Mixture Coefficient $\delta = [0, 0.2, 0.4, 0.6, 0.8, 1.0]$ are evaluated by increasing the dependency to the labels. Each value of $\delta$ determines a feature selection model. Then the resulting kernel $K_{\delta}$ is used for support vector classification and evaluated on the test set.\\ Figure 2 shows the AUC results on classification by using features selected for different values of $\delta$ on the three datasets. The AUC score peaks between $\delta = 0.4$ and $\delta = 0.6$ in all the cases. This results evidence the importance of the latent regularization in classification and means that the maximum AUC score using the KLR-FS features is obtained with a mixture between supervised labels and unsupervised latent structure.\\
Figure 3 shows how the RED score varies for different values of the Mixture Coefficient $\delta$ on the corresponding selected features and how the RED score is reduced by the latent regularization. For the BRCA-US and PACA-CA datasets the lowest values of RED are obtained for $\delta = 0.4$ and $\delta = 0.6$ and match with the same range of $\delta$ values with maximum AUC. For the SMK-CAN dataset the lowest RED scores are obtained between $\delta = 0.6$ and $\delta = 1$ and only when the number of selected features $p = 20$ and $p = 30$ the lowest RED score is at $\delta = 1$. Figure 3 shows that when the number of selected features $p$ is small the mixture parameter $\delta$ has a stronger impact.

\begin{figure}[H]
  \centering 
  \includegraphics[width=3.4in]{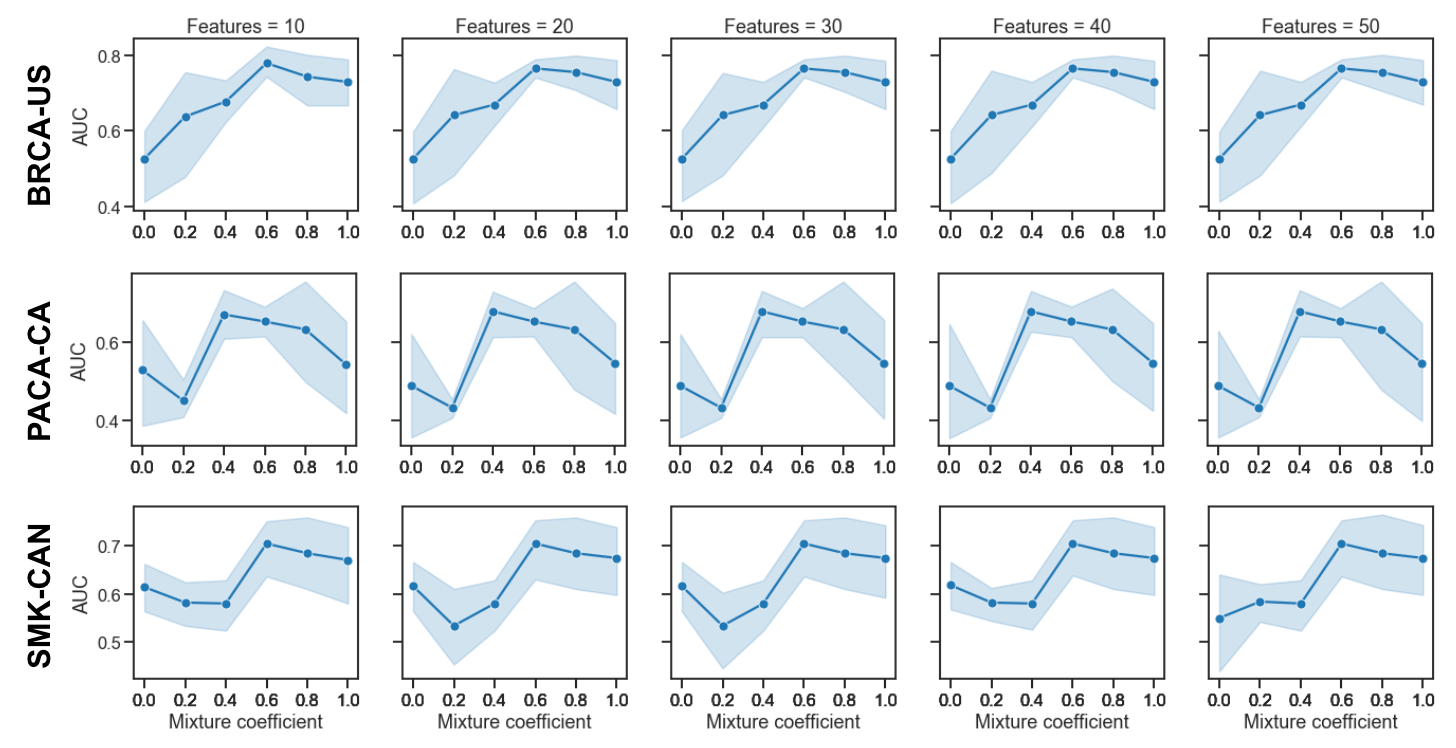} 
  \vspace*{-3mm}
  \caption{Classification performance measured by AUC-ROC using different number of selected features by KLR-FS across different values of the $\delta$ mixture coefficient.}
  \label{fig:example} 
\end{figure} 
\begin{figure}[H]
  \centering 
  \includegraphics[width=3.4in]{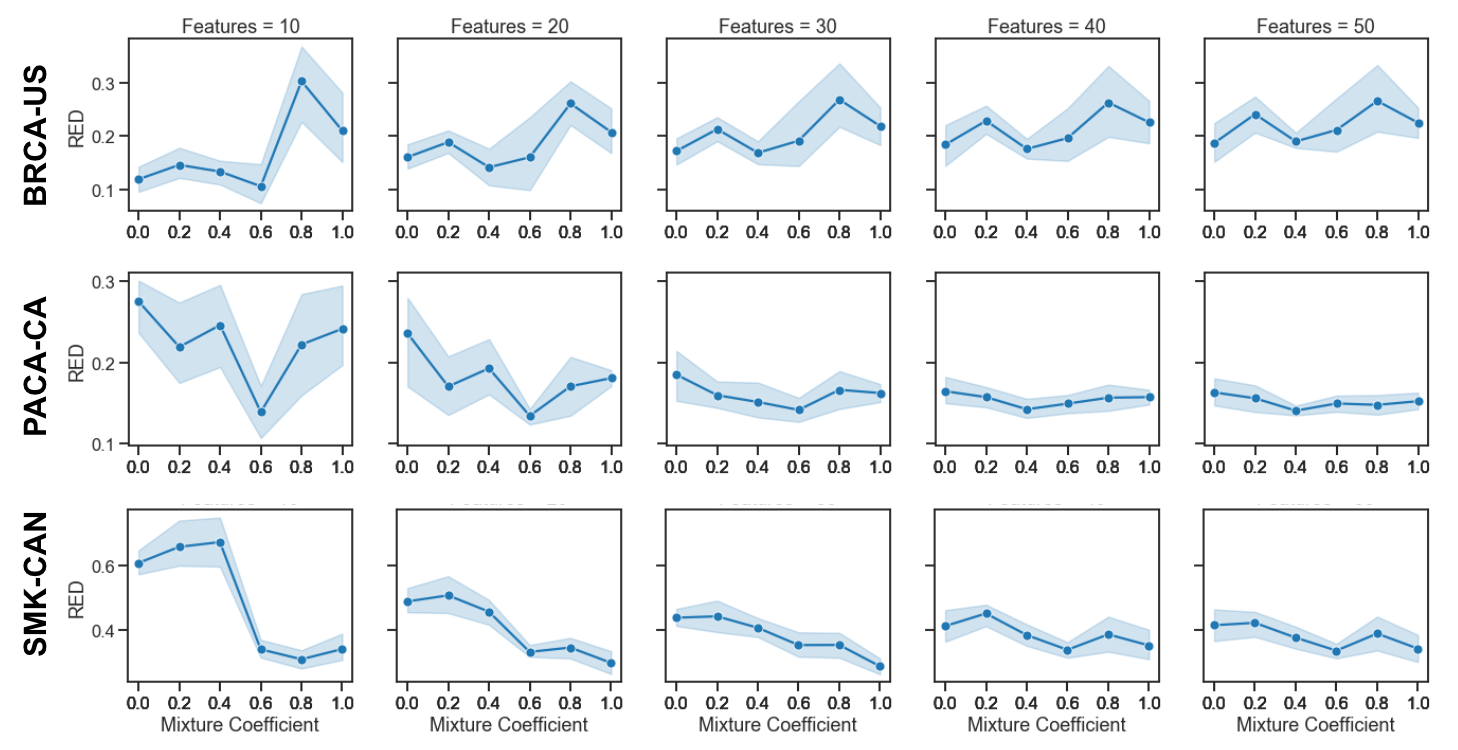} 
  \vspace*{-3mm}
  \caption{Redundancy rate RED of the selected features by KLR-FS with different values of the $\delta$ mixture coefficient.}
  \label{fig:example} 
\end{figure} 

\subsection{Performance estimation and statistical comparison}

In this subsection the KLR-FS is compared with the benchmark methods for different number of selected features $p$. For performance estimation the results of the classification and selection tasks are averaged among all the random iterations. The feature selection is applied on train samples and the classifier is trained on the selected features to finaly classify the test samples. The mixture coefficient used for KLR-FS for benchmark is $\delta = 0.6$ since it appear to be the best combination according to AUC and RED scores in the previous section.

\begin{figure}[H]
  \centering 
  \includegraphics[width=3.4in]{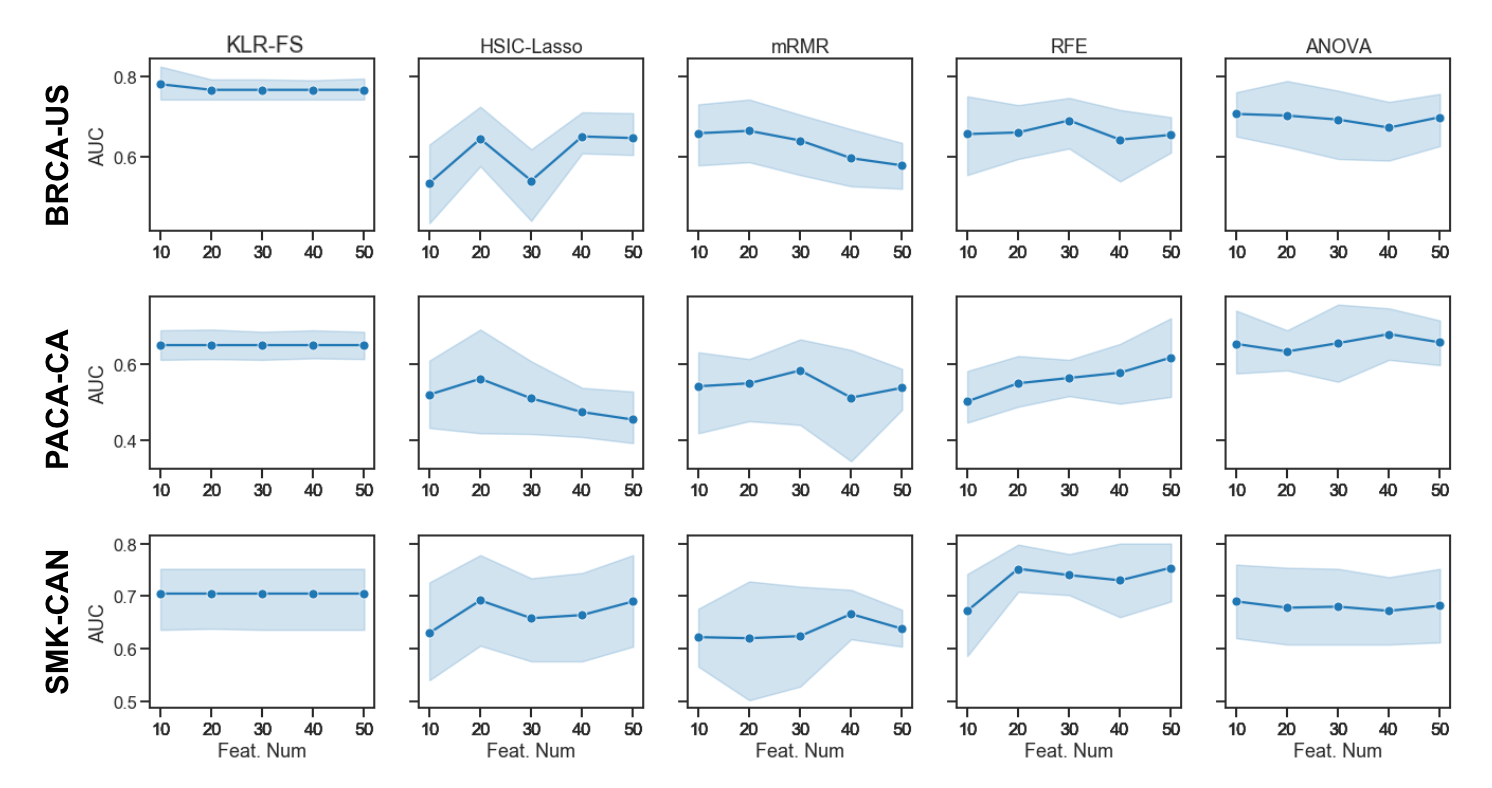} 
  \vspace*{-4mm}
  \caption{Comparison of the classification performance measured by AUC-ROC between KLR-FS and benchmark methods for different number of selected features $p$ and different datasets.}
  \label{fig:example} 
\end{figure} 
Figure 4 shows the classification performance on test set for different numbers of $p$ selected features by each method. For the BRCA-US and PACA-CA datasets the KLR-FS shows the highest mean AUC score and the lowest classification variance for every number of selected features. For the SMK-CAN dataset the KLR-FS has the highest classification performance with $p = 10$ features and for larger number of features it is outperformed by the RFE method. \\
In the three datasets the classification results using the KLR-FS features are the most constant as the value of $p$ increases. According to section 3.1 the resulting $K_{\mu}$ assigns a larger weight $\mu_i$ to the variables that are selected first since these are the ones that increments more the overall alignment $\Delta A(K_{\delta}, K_{\mu})$, while to those variables selected later the corresponding weight is much smaller. The KLR-FS results on Figure 4 suggest that these classification problems can be solved by selecting the first 10 variables since selecting more variables does not introduce a significant improvement in the classification results.
\begin{figure}[H]
  \centering 
  \includegraphics[width=3.4in]{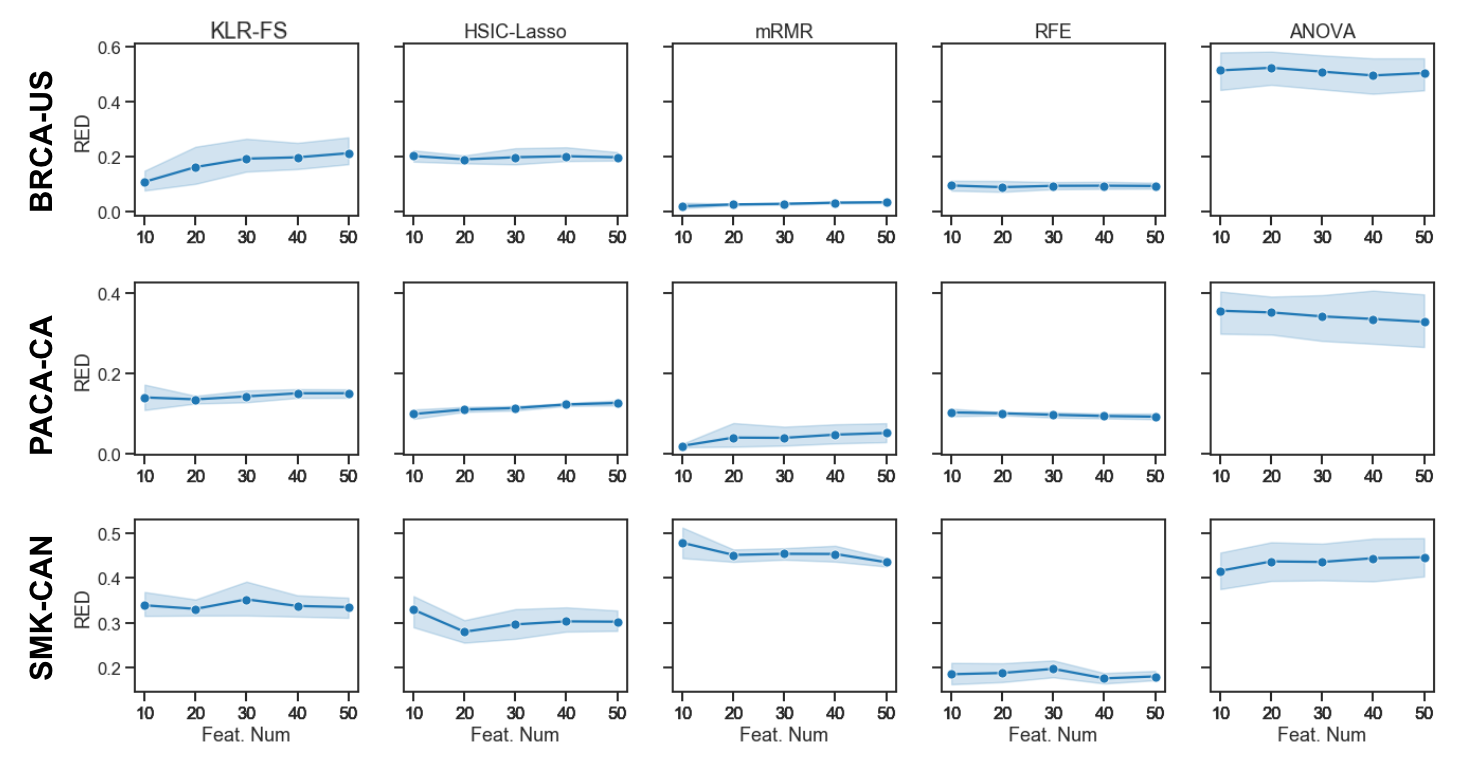} 
  \vspace*{-4mm}
  \caption{Evolution of the RED score for different number of features on each method.}
  \label{fig:example} 
\end{figure} 
Figure 5 shows the RED score of each subset of selected features for different values of $p$. For the BRCA-US and PACA-CA datasets the lowest score is obtained by the mRMR method with a $RED < 0.1$ followed by RFE, KLR-FS and HSIC-Lasso. For the SMK-CAN dataset the lowest RED score is obtained by the RFE method followed by HSIC-Lasso and KLR-FS while the highest corresponds to mRMR and ANOVA. The ANOVA shows the highest RED score in all datasets. KRL-FS and HSIC-Lasso shows a similar behaviour with a $RED < 0.2$ in the BRCA-US and PACA-CA datasets and a $RED < 0.4$ for the SMK-CAN datasets. \\
This section compares the KLR-FS method with $\delta = 0.6$ and four feature selection methods. The main objective of this work is to select features to improve the tumor classification. It is observed that the classification performance of the KLR-FS is the highest for the majority of the cases.

\section{Discussion}
This work proposes a feature selection model based on Multiple Kernel Learning coupled with kernel-PCA with an application to tumor classification using high dimensional gene expression data from tumor profiles. The proposed method KLR-FS aims to select features considering not only the sample labels $y$ but also the latent structure $z$ of the training data presented in this work as a \textit{latent regularization} since it is used to improve the generalization capacity in feature selection for classification. The idea is based on considering not only a pure supervised target kernel $k_{yy}$ but also a kernel $k_z$ built from the latent variables $z$ obtained from a nonlinear dimensionality reduction $\phi_z$. The latent regularization comes from a kernel built on the latent space $\mathcal{Z}$ generated by the kernel-PCA. This allows to explore the relaxation of the sample labels by a linear combination of a supervised and an unsupervised kernel. The experiments detailed on section 5.1 show that the highest classification performance and the lowest redundancy rate for the KLR-FS is obtained when latent information is mixed with label targetted information. These results show that selecting variables contemplating both the latent structure $z$ of the data and the labels of the supervised problem can improve the classification task. Moreover, learning partially a latent space works as a regularization term since it limit the solution space by introducing the need to capture also the general structure of the data. By doing so it can improve the generalization capacity on new unseen test samples. In addition, a relation between the latent structure of the data and the sample labels $y$ exists and only learning from the labels with a classic supervised criteria can result in overfitting when training in a context where the number of samples is very limited. Figure 2 reveals how the latent regularization works as a relaxation of the supervised problem by mixing $K_{yy}$ with $K_{z}$. This evidences the important role of the unsupervised latent variables in the supervised feature selection task. As expected, a value of $\delta = 0$ almost does not select any useful feature since the classification performance declines significantly in almost all cases. Finally, there is consistency between the peak in the AUC and the lowest RED score in the majority of the cases with a mixture coefficient ranging between $0.4 < \delta < 0.6$.\\ 
Despite the kernel-PCA is used as a nonlinear and unsupervised mapping function $\phi_z(x)$ to learn a low dimensional latent space from training data any other nonlinear transformation could be used like Autoencoders or t-SNE \cite{palazzo2019pan} \cite{kobak2019art}. In this work since the tumor data has a considerably small sample set in comparison to the number of features $m<<n$ then Autoencoders and t-SNE were not considered although they could be used in cases where more samples are available.\\
KLR-FS provides feature importance via the non-zero elements of the resulting sparse $\mu$ vector, which is a useful characteristic to gain interpretation of the results. The model shows that it can deal with high dimensional initial problems, in this case the initial dimension in all datasets is $n>18.000$ and low sample sizes where the classification using the KLR-FS features outperforms the benchmark methods. Despite the KLR-FS is only outperformed in RED score by RFE in the three datasets and by mRMR in two datasets, KLR-FS has the highest classification performance and the lowest variance when compared with all the benchmark methods. \\ KLR-FS outputs a custom kernel $K_{\mu}$ that is used in support vector classification and contributes to improve the classification performance. The resulting $K_{\mu}$ kernel can be used not only for classification tasks but also for representation of tumor profiles or visualization by kernel-PCA since it is built from both supervised and unsupervised latent sources. \\
The proposed method has relevance in problems where the structure of the training data correlates partially with the tumor labels thus learning also from the latent structure of the data improves the supervised learning task.

\section{Conclusion}

In a context of low sample size and high dimensional space we propose a novel feature selection method that want to enforce the selected features to be discriminant and to keep information about the general data structure. By doing so we expect to avoid overfitting, to improve generalization and finally classification performance. To reach that goal we design a new approach called \textit{Kernel Latent Regularization Feature Selection} (KLR-FS). It is based on a MKL approach that targets a label kernel relaxed with another kernel built on a latent space. In the proposing application the latent space is obtained using kPCA. The method is applied on high dimensional gene expression tumor profiles from Breast, Pancreas and Lung Cancer. KLR-FS selects genes which are used to classify tumor subtypes or survival rate with the highest classification performance and a considerably low redundancy rate. The mixture between supervised labels and latent variables reveals an improvement in the generalization capacity when compared with other feature selection methods. \\ 
Future work should generalize to multi-modal data as multi-omic layers of biological information to not only select features but also improve the fusion of heterogeneous data.


\section*{Acknowledgements}
We thank to Diego Tomassi and Ivan Lengyel for helpful comments on an early version of this work.

\section*{Funding}
This work was supported by the Universite de Technologie de Troyes in France and the Universidad Tecnologica Nacional in Argentina. Also it was funded by grants from CONICET, ANPCyT and FOCEM-Mercosur.

\bibliography{article_ref}

\end{document}